\documentclass[10pt,twocolumn,letterpaper]{article}
\usepackage{titlesec}
\titlespacing*{\paragraph} {0pt}{1pt}{2pt}
\titlespacing*{\section} {0pt}{1pt}{1pt}
\titlespacing*{\subsection} {0pt}{1pt}{1pt}

\usepackage{cvpr}              

\usepackage{graphicx}
\usepackage{amsmath}
\usepackage{amssymb}
\usepackage{booktabs}
\usepackage{multicol}
\usepackage{multirow}
\usepackage{xcolor, soul}
\usepackage{bbm}
\usepackage{xr-hyper} 

\newcommand{\point}[1]{\noindent\textbf{#1}}
\DeclareMathOperator*{\argmax}{arg\,max}

\newcommand{\ginanote}[1]{}
\newcommand{\highlight}[1]{}
\newcommand{\ar}[1]{}
\newcommand{\comment}[1]{}
\newcommand{\zl}[1]{}
\newcommand{\guru}[1]{}
\renewcommand{\st}[1]{}

\makeatletter
\newcommand*{\addFileDependency}[1]{
  \typeout{(#1)}
  \@addtofilelist{#1}
  \IfFileExists{#1}{}{\typeout{No file #1.}}
}
\makeatother

\usepackage[pagebackref,breaklinks,colorlinks]{hyperref}

\usepackage[capitalize]{cleveref}
\crefname{section}{Sec.}{Secs.}
\Crefname{section}{Section}{Sections}
\Crefname{table}{Table}{Tables}
\crefname{table}{Tab.}{Tabs.}

\def\cvprPaperID{676} 
\def\confName{CVPR}
\def\confYear{2022}

\begin{document}

\title{Class-Incremental Learning with Strong Pre-trained Models} 
\author{Tz-Ying Wu$^{1,2}$ \quad Gurumurthy Swaminathan$^1$ \quad Zhizhong Li$^1$ \\ Avinash Ravichandran$^1$ \quad Nuno Vasconcelos$^2$ \quad Rahul Bhotika$^1$ \quad Stefano Soatto$^1$\\[-1.5em]
\and
$^1$AWS AI Labs\\
{\tt\small\{gurumurs,lzhizhon,ravinash,bhotikar,soattos\}@amazon.com}\\
\and
$^2$UC San Diego\\
{\tt\small\{tzw001,nuno\}@ucsd.edu}
}
\maketitle

\begin{abstract}
Class-incremental learning (CIL) has been widely studied under the setting of starting from a small number of classes (base classes). Instead, we explore an understudied real-world setting of CIL that starts with a strong model pre-trained on a large number of base classes. We hypothesize that a strong base model can provide a good representation for novel classes and incremental learning can be done with small adaptations. We propose a 2-stage training scheme, i) feature augmentation -- cloning part of the backbone and fine-tuning it on the novel data, and ii) fusion -- combining the base and novel classifiers into a unified classifier. Experiments show that the proposed method significantly outperforms state-of-the-art CIL methods on the large-scale ImageNet dataset (\eg +10\% overall accuracy than the best). We also propose and analyze understudied practical CIL scenarios, such as base-novel overlap with distribution shift. 
Our proposed method is robust and generalizes to all analyzed CIL settings.
\end{abstract}

\section{Introduction}\label{sec:intro}

As deep classifiers become more popular for real-world applications, the need for incrementally learning novel classes (novel data) becomes more prevalent. Training a classifier with both old and novel data is not optimal when old data can become unavailable over time~\cite{ewc,si,lwf,icarl,eeil,rusu2016progressive,packnet,piggyback}. Fewer old data leads to a high imbalance between the old and novel data, and simply fine-tuning the model causes catastrophic forgetting for the old classes~\cite{lwf}.

\begin{figure}
    \centering
    \includegraphics[width=\linewidth,height=0.6\linewidth]{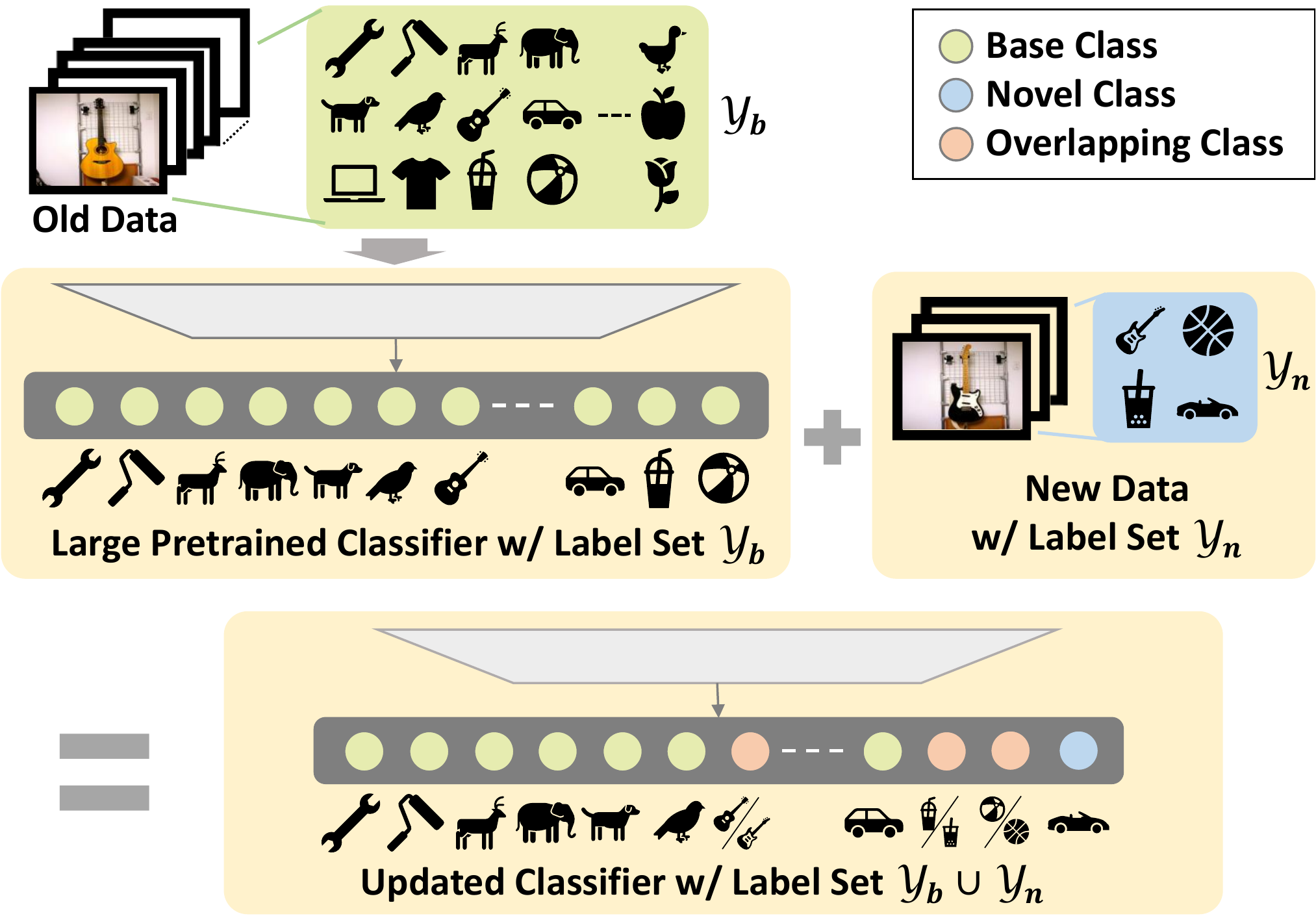}
    \caption{We study the problem of CIL in the setting where there are a large number of base classes. Classes between splits can have overlaps, and data can be sampled from the same or different distributions (\eg different styles, poses).}
    \label{fig:teaser}
\end{figure}

Class-incremental learning (CIL) methods~\cite{icarl,eeil,bic,podnet,lucir,wa,rainbowmem} learn to categorize more and more classes over time. However, they typically start their incremental training with a small number of base classes (\eg only 50), and add an equally small number of new classes at a time. In many practical scenarios, having a large number of base classes can be a more useful starting point for building an application. For example, a strong model could have been developed to identify different dog breeds, and a small set of additional breeds needs to be added for model update. Moreover, base and novel classes may overlap but may distribute differently, such as a guitar class present in both base and novel classes, with only acoustic guitars in base samples while electric guitars in novel ones.

Some CIL methods use a static model and typically fine-tune the existing parameters with some constraints imposed on parameter changes~\cite{ewc,si,mas}, gradients~\cite{gem}, features~\cite{lucir,podnet}, or activations~\cite{lwf,icarl,eeil}. These methods  modify the well-trained network weights and risk performance degradation. On the other hand, methods based on dynamic models learn separate parameters for novel tasks, either by expanding the model~\cite{rusu2016progressive}, or introducing a parameter gating function~\cite{packnet,piggyback}. However, almost all dynamic methods focus on task-incremental learning (TIL). TIL assumes that which task a sample belongs to is known at inference time, and different tasks are inferred individually. This assumption is not realistic if the application needs to distinguish between base and novel classes. A recent work DER~\cite{der} uses a dynamic model for CIL, where it duplicates the entire backbone for novel data and prunes the model.

Current CIL approaches may not be optimal when a large number of base classes (\eg 800) is used to pre-train a strong model. We hypothesize that the well-trained backbone is capable of extracting representative features for the novel data and freezing it partially while learning a small adaptation branch for novel data works better than fine-tuning the whole backbone. We show in a preliminary study that fine-tuning fewer layer blocks outperforms full fine-tuning when using a strong pre-trained model.

Hence, we propose a 2-stage training scheme for CIL starting with a large number of base classes: i) duplicating part of the backbone as the adaptation module and fine-tuning it on the novel data, and ii) combining all the independently trained base and novel classifiers into a unified classifier at each incremental step. Towards this, we propose a score fusion network that enables  knowledge transfer between base and novel classes by combining the logits. While the optimal adaptation module size may depend on the old and novel data discrepancy, we show that our score fusion generalizes to different adaptation module sizes. 

Most CIL research~\cite{icarl,eeil,bic,podnet,lucir,wa} only consider the scenario where the base and novel label sets are disjoint. Rainbow Memory~\cite{rainbowmem} (RM) is the only exception where label sets are identical among tasks but class frequencies differ. In this work, we explore a more general setting where some novel classes can overlap with base classes, potentially with a different distribution (\eg, different styles, poses), as shown in Figure~\ref{fig:teaser}. We show how our score fusion can handle overlapping classes by using a knowledge pooler to combine their base and novel logits.

In summary, we provide three contributions: 


\begin{itemize}
\setlength\itemsep{0mm}
    \item We propose a 2-stage CIL training strategy. In stage-I, instead of tuning the base network and risking catastrophic forgetting, we replicate part of the network and fine-tune the extra branch on novel data. We show that as we start with a strong pre-trained base network, this approach outperforms state-of-the-art CIL methods.
    \item We propose a new score fusion algorithm for stage-II, where we unify the classifiers for base and novel data into one by consolidating their output logits.
    \item We generalize CIL to a broader and more challenging scenario: base and novel classes can partially overlap, and the overlap classes may have changed distributions (\eg new style). We show that our method is robust and generalizes to this new scenario.
\end{itemize}

\section{Related work}
Prior work in continual learning 
can be mainly categorized into two streams: methods using a (1) static and (2) dynamic model architecture. For the first group, the general approach for incremental training is to impose constraints on parameter changes~\cite{ewc,si,mas}, gradients~\cite{gem}, features~\cite{lucir,podnet,tpcil}, or activations~\cite{lwf,icarl,eeil,hou2018lifelong,ssil} when fine-tuning the network with novel samples. They can be either memory-free~\cite{ewc,si,lwf,mas} which only rely on the novel data and the original point of convergence in parameters, or be memory-based~\cite{icarl,gem,ewc,eeil,lucir,podnet,bic,wa,ssil} which relaxes the constraint and keep exemplars of old data for replay. The exemplars can be selected in many ways, such as herding~\cite{icarl}, random sampling~\cite{hou2018lifelong}, uncertainty sampling~\cite{rainbowmem}. 


The second line of work adopt dynamic models, where task-specific parameters are introduced to prevent interference among tasks, which can be achieved by growing the model~\cite{rusu2016progressive,growabrain,zhang2020side,dan,der,li2019learn,den,rebuffi2017learning,nettailor}, or introducing a gating function on the parameters~\cite{packnet,piggyback,channelgate}. However, most works in this line either focus only on the performance of the target task~\cite{growabrain,dan,rebuffi2017learning,nettailor}, or assume task label is available at inference time~\cite{rusu2016progressive,zhang2020side,packnet,piggyback,cpg,li2019learn,den}. The exceptions infer the task identity typically by learning a routing classifier~\cite{expertgate,channelgate}, or estimating confidence scores~\cite{supsup}. However, they do not produce a unified head for all the classes.
 
Our proposal is a two-stage method in the offline setting with a dynamic model. In stage-I, we expand the network by cloning and fine-tuning partial parameters from the original backbone. While DER~\cite{der} shares some commonalities with our first stage, we differ in stage-II. We show the downside of retraining a linear classifier for all classes solely with the data in the memory as in~\cite{der,rainbowmem}, and propose to combine the logits of the old and new expert classifiers. We also investigate consolidating knowledge of overlapping classes, which is understudied in the literature.  

Our method bears superficial similarity to transfer learning and few-shot learning in that it uses a large number of base classes. However, they focus only on the performance of novel classes. Such methods underperform on overall average performance in our experiments. 


\section{CIL with strong pre-trained models}
In this section, we present the problem formulation and the details of the proposed method.


\subsection{Problem formulation}
Given a dataset $\mathcal{D}=\{(\mathbf{x}_i, y_i)\}_{i=1}^{N}$, where $x_i$ and $y_i$ are data and label respectively, the goal of a traditional classification network is to learn a feature extractor $h(\cdot;\mathbf{\Phi})\in\mathbb{R}^{k}$ and a linear classifier  $\mathbf{W}\in\mathbb{R}^{k\times|\mathcal{Y}|}$, where $\mathcal{Y}$ is the label set of $\mathcal{D}$. This is usually obtained by minimizing the cross-entropy loss,
\begin{align}
    &L_{ce} = -\frac{1}{N}\sum_{i=1}^N \log \hat{p}^{(y_i)}(\mathbf{x}_i) \label{eq:ce}
    \end{align}
where
\begin{align}
     \hat{p}(\mathbf{x}) = \sigma(\mathbf{W}^T h(\mathbf{x}; \mathbf{\Phi})),
\end{align}
$\sigma(\cdot)$ is the softmax function, and $\mathbf{v}^{(l)}$ is the $l^{\text{th}}$ element of $\mathbf{v}$.  

In class-incremental learning (CIL), we first focus on one incremental step, and generalize into multiple steps in Section~\ref{sec:multistep}. Given a base model $\mathcal{M}_b$ pre-trained on a label set $\mathcal{Y}_b$ using the base dataset $\mathcal{D}_b$,  we augment $\mathcal{Y}_b$ with another label set $\mathcal{Y}_n$ using dataset $\mathcal{D}_n$, \ie, the new all-labels set $\mathcal{Y}_a = \mathcal{Y}_b \cup \mathcal{Y}_n$.
Most prior works~\cite{icarl,eeil,bic,podnet,lucir,wa} focus on the CIL scenario where the label sets are fully disjoint, \ie $\mathcal{Y}_b\cap \mathcal{Y}_n=\emptyset$, while in practice, classes can overlap between base and novel splits~\cite{rainbowmem}, where $|\mathcal{Y}_a|<|\mathcal{Y}_b|+|\mathcal{Y}_n|$. A full-overlapping extreme case is $\mathcal{Y}_n\subseteq \mathcal{Y}_b$, which means the new label set $\mathcal{Y}_a = \mathcal{Y}_b$. Despite the overlap, base and novel samples can be very different in each of the overlapping classes, such as having different poses or styles. 

An intuitive solution to learn the combined label set is to use $\mathcal{D}_b \cup \mathcal{D}_n$ to train a standard classifier with eq.~(\ref{eq:ce}). However, as motivated in the CIL literature, $\mathcal{D}_b$ often becomes unavailable over time in practice. In this case, $\mathcal{D}_n$ dominates the training set and causes
catastrophic forgetting, degrading base class performance significantly.

Traditional CIL methods divide existing datasets' classes evenly into multiple splits, each with a small number of classes (\eg 50). Typically, the initial model is briefly trained only on one of the small splits. Instead, we explore a real-world scenario that is understudied in the literature, where the pre-trained model has been sufficiently trained on a large number of classes (\eg 800 classes). In following incremental steps $|\mathcal{Y}_b| \gg |\mathcal{Y}_{n}|$, so the smaller novel splits bring comparably little additional information. 

\subsection{Can minimal network modification work well?}
If we freeze the pre-trained network, we avoid catastrophic forgetting. However, this representation might not generalize to the novel classes, especially in traditional CIL where the first-step model is pre-trained with a small number of classes. To mitigate this issue,  most prior CIL work~\cite{icarl,eeil,bic,podnet,lucir,wa} update the feature extractor with novel data. Regularization in the loss function (such as penalizing changes in predictions~\cite{lwf,icarl} or network weights~\cite{ewc,si}, or re-training on saved examplars~\cite{bic,wa}) can reduce catastrophic forgetting to a certain extent, but it still remains prevalent and unsolved.

In this paper, we propose to add extra capacity to the network. We differ from prior network-growing approaches~\cite{rusu2016progressive,growabrain} in two ways. 
First, our goal is CIL rather than TIL, \ie, the network must avoid confusion between novel and base classes. This is not solved by merely freezing existing network weights, which by construct only avoids forgetting in classifying among old classes.
More importantly, CIL with a large number of base classes allows our method to leverage the powerful pre-trained model. We hypothesize that drastic changes in the existing network is unnecessary, and modifying only the top few layers will suffice. Before we introduce our method, we validate this hypothesis and our strategy of minimal network changes with two motivational experiments.



\begin{figure}
    \centering
    \includegraphics[width=0.49\linewidth]{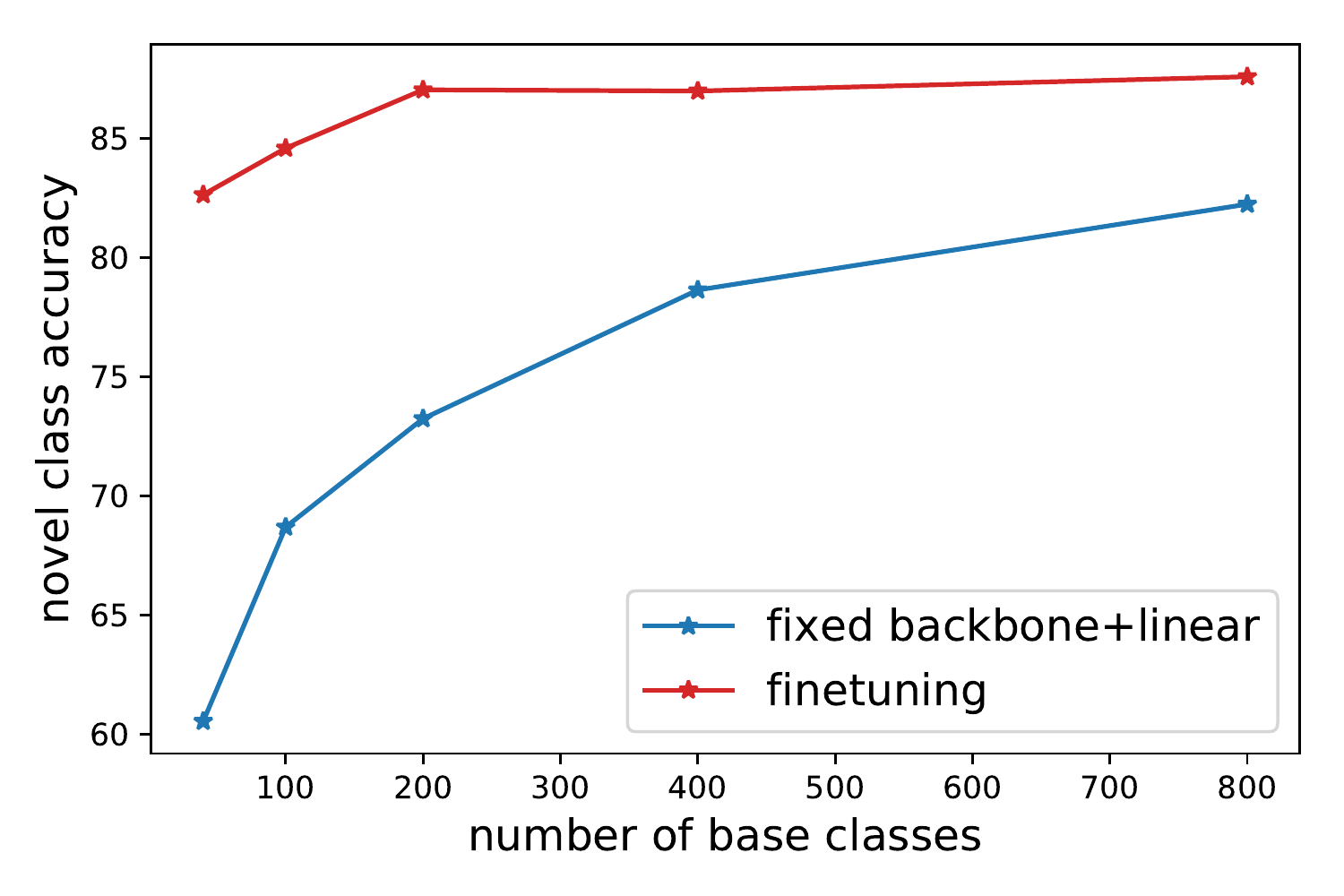}
    \includegraphics[width=0.49\linewidth]{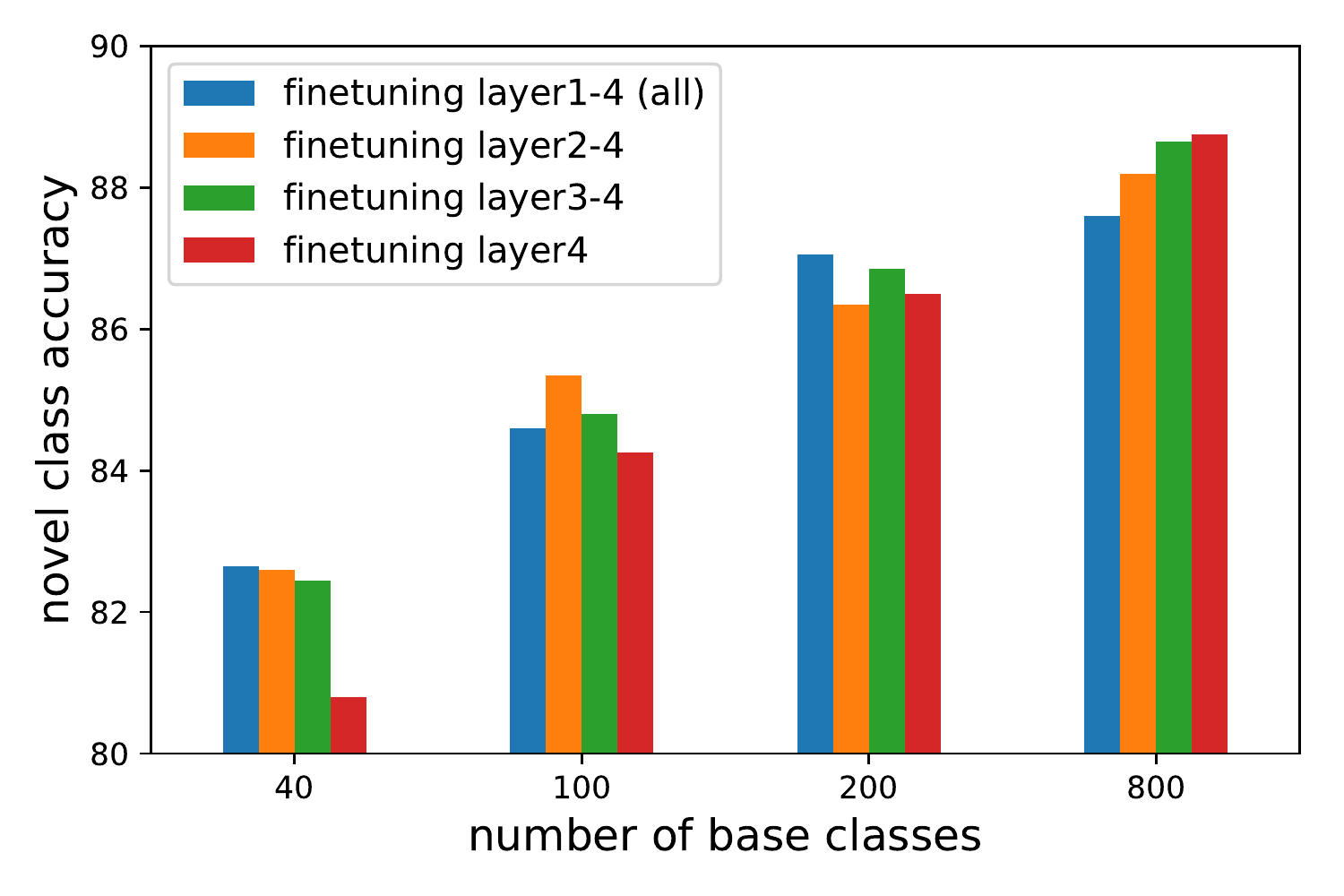}
    \caption{Left: The comparison between fine-tuning and freezing representations. Right: The novel class accuracy of fine-tuning each pre-trained model with different numbers of layers.}
    \label{fig:prelim}
\end{figure}

We first analyze {\it when there are many base classes, how much (or little) benefit is in optimizing representations with the novel data}. Starting from a pre-trained ResNet10~\cite{he2016deep} as $\mathcal{M}_b$, we train a classifier for $\mathcal{Y}_n$ only using new data $\mathcal{D}_n$, with or without fine-tuning $h(\cdot;\mathbf{\Phi})$. As shown in Figure~\ref{fig:prelim} (left), the gap in novel class accuracy between fine-tuning and fixing the feature extractor is large when the number of base classes is small. The gap significantly reduces with increasing number of base classes, but does not disappear even with $|\mathcal{Y}_b|=800$. This indicates that fully changing the representations for the novel data is still beneficial for learning $\mathcal{D}_n$, but with greatly reduced significance. 

We then explore {\it whether we need to optimize the entire pre-trained feature extractor using novel data}.  Figure~\ref{fig:prelim} (right) presents the novel class accuracy of fine-tuning smaller subsets of the four convolutional blocks in ResNet10. With weaker pre-trained models (\eg $|\mathcal{Y}_b|=40,100,200$), almost all the layers in the backbone need to be fine-tuned to gain good performance in novel classes. However, for strong pre-trained models (\ie, $|\mathcal{Y}_b|=800$), fine-tuning anything beyond the last convolutional block (\ie, layer4) degrades the strong performance. 

\subsection{Training pipeline}
\begin{figure}
    \centering
    \includegraphics[width=0.7\linewidth]{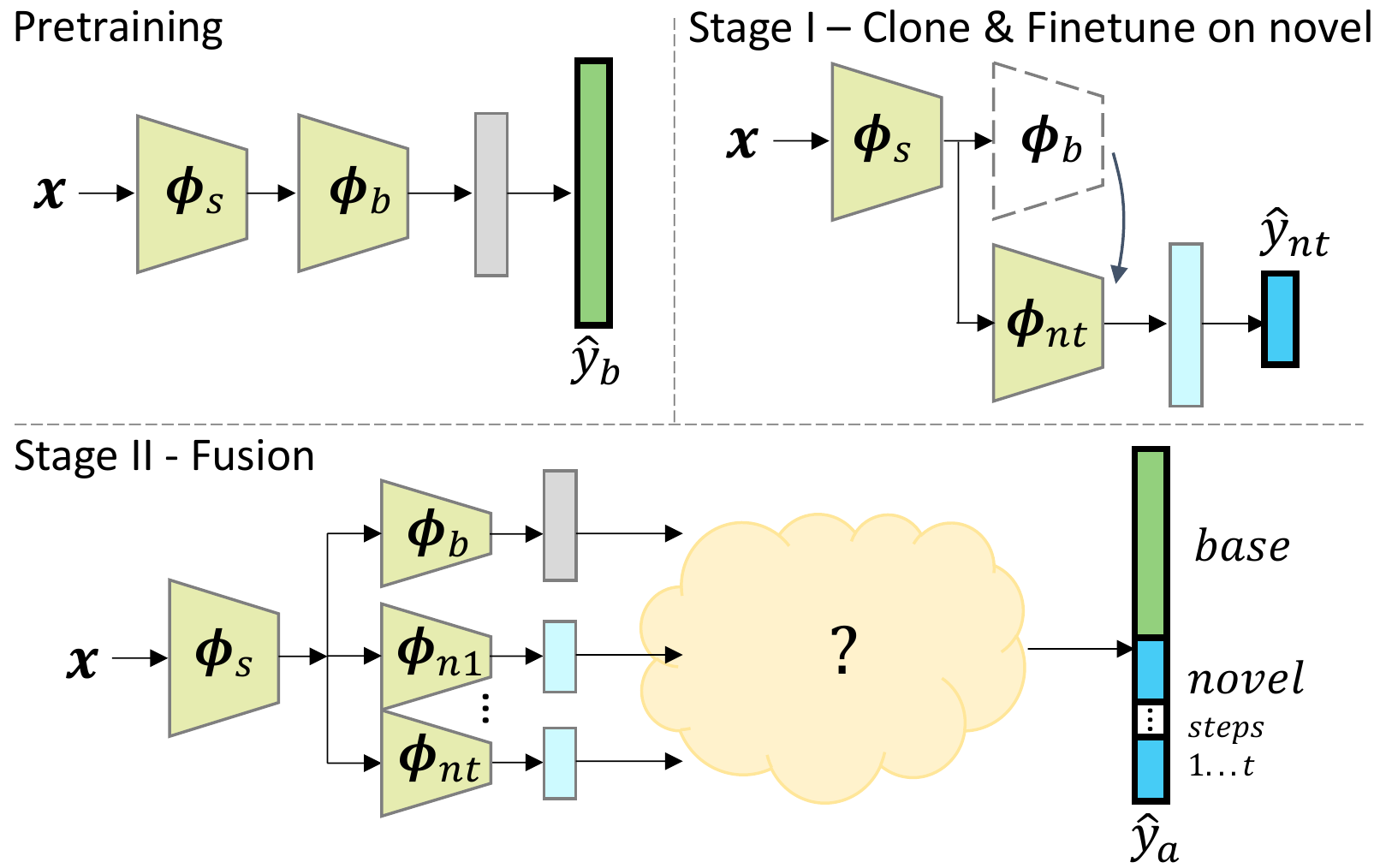}
    \caption{General training pipeline.}
    \label{fig:stages}
\end{figure}

Inspired by these observations, we propose a 2-stage incremental training pipeline (Figure~\ref{fig:stages}).  We first formulate using one incremental step (base+novel):

\point{Stage-I -- Feature augmentation (FA)}: We split the pre-trained feature extractor $h(\cdot;\mathbf{\Phi})$ into two sub-networks, where the encoder with parameter $\mathbf{\Phi}_s$ is followed by the encoder with parameter $\mathbf{\Phi}_b$, $\mathbf{\Phi}=\{\mathbf{\Phi}_s, \mathbf{\Phi}_b\}$. To optimize the features for $\mathcal{D}_n$ without forgetting the ones for $\mathcal{D}_b$, we expand the features by cloning the sub-network of $\mathbf{\Phi}_b$ to the branch $\mathbf{\Phi}_n$ as the adaptation module, and fine-tuning $\mathbf{\Phi}_n$ and the weights of the novel class classifier $\mathbf{W}_n\in\mathbb{R}^{k\times|\mathcal{Y}_n|}$ on $\mathcal{D}_n$ with the loss of eq.~(\ref{eq:ce}). The shared $\mathbf{\Phi}_s$ is frozen. This setup ensures no forgetting in the old representation, while enabling feature learning to accommodate new knowledge.  
While the optimal size of the frozen $\mathbf{\Phi}_s$ depends on the data discrepancy between base and novel splits, which itself can be a research topic to explore, we adopt the last convolutional block (\ie layer4) as $\mathbf{\Phi}_b$ as it suffices for most cases, and $\mathbf{\Phi}_s$ is the layer1-3 blocks.

\point{Stage-II -- Fusion}: After the first-stage training, we have the base and novel classifiers, $\mathcal{M}_b(\cdot;\mathbf{\Phi}_s,\mathbf{\Phi}_b,\mathbf{W}_b)$ and $\mathcal{M}_n(\cdot;\mathbf{\Phi}_s,\mathbf{\Phi}_n,\mathbf{W}_n)$, optimized for $\mathcal{Y}_b$ and $\mathcal{Y}_n$ respectively. We introduce in Section~\ref{sec:fusion} our score fusion scheme to combine the knowledge of the two networks and get a unified classifier of the label set $\mathcal{Y}_a$, where $\mathcal{Y}_a=\mathcal{Y}_b\cup \mathcal{Y}_n$. 

\label{sec:multistep}
\point{Multiple incremental steps}: We maintain one extra branch per new step: $\{\mathbf{\Phi}_b, \mathbf{W}_b, \mathbf{\Phi}_{n1}, \mathbf{W}_{n1}, \dots, \mathbf{\Phi}_{nT}, \mathbf{W}_{nT}\}$. In each novel step $t\in\{1..T\}$, we initialize the new branch $\mathbf{\Phi}_{nt}$ with $\mathbf{\Phi}_b$, fine-tune it, and re-fuse all $t+1$ branches.



\begin{figure*}
    \centering
    \begin{minipage}[t]{\linewidth}
    \centering
    \begin{minipage}[h]{0.224\linewidth}
        \centering
        \includegraphics[width=\linewidth]{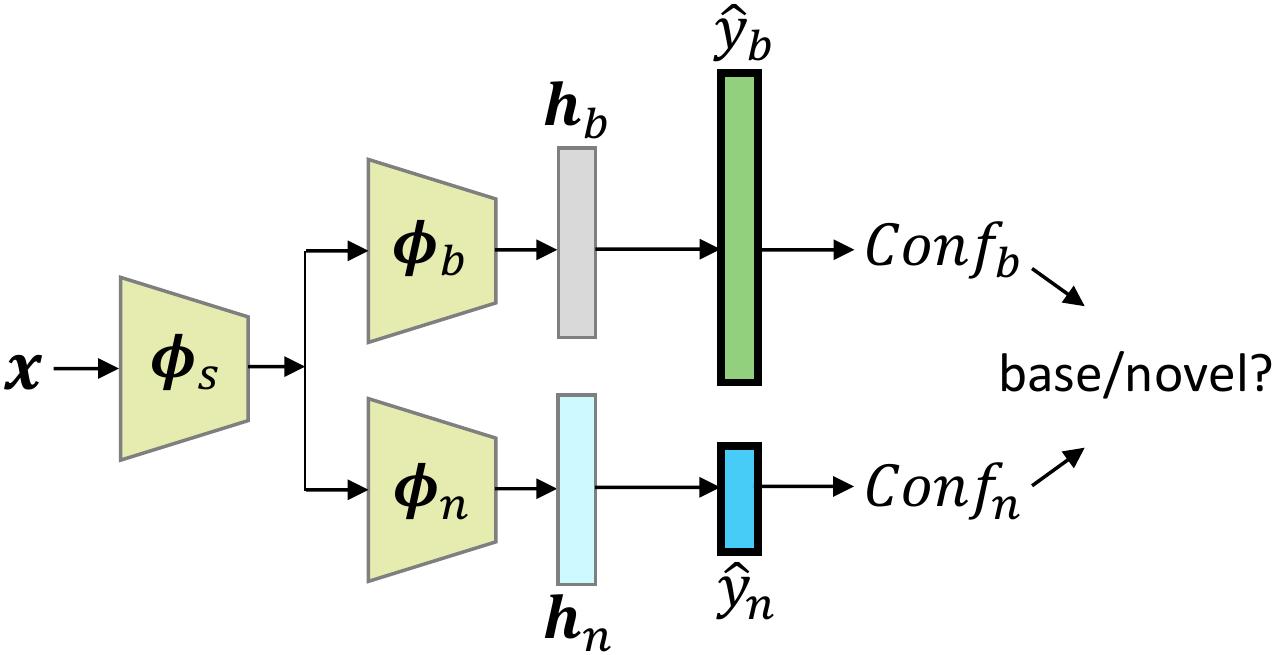}
        \subcaption{\footnotesize Confidence-based routing baseline.}
        \label{fig:route_conv}
    \end{minipage}
    \begin{minipage}[h]{0.224\linewidth}
        \centering
        \includegraphics[width=\linewidth]{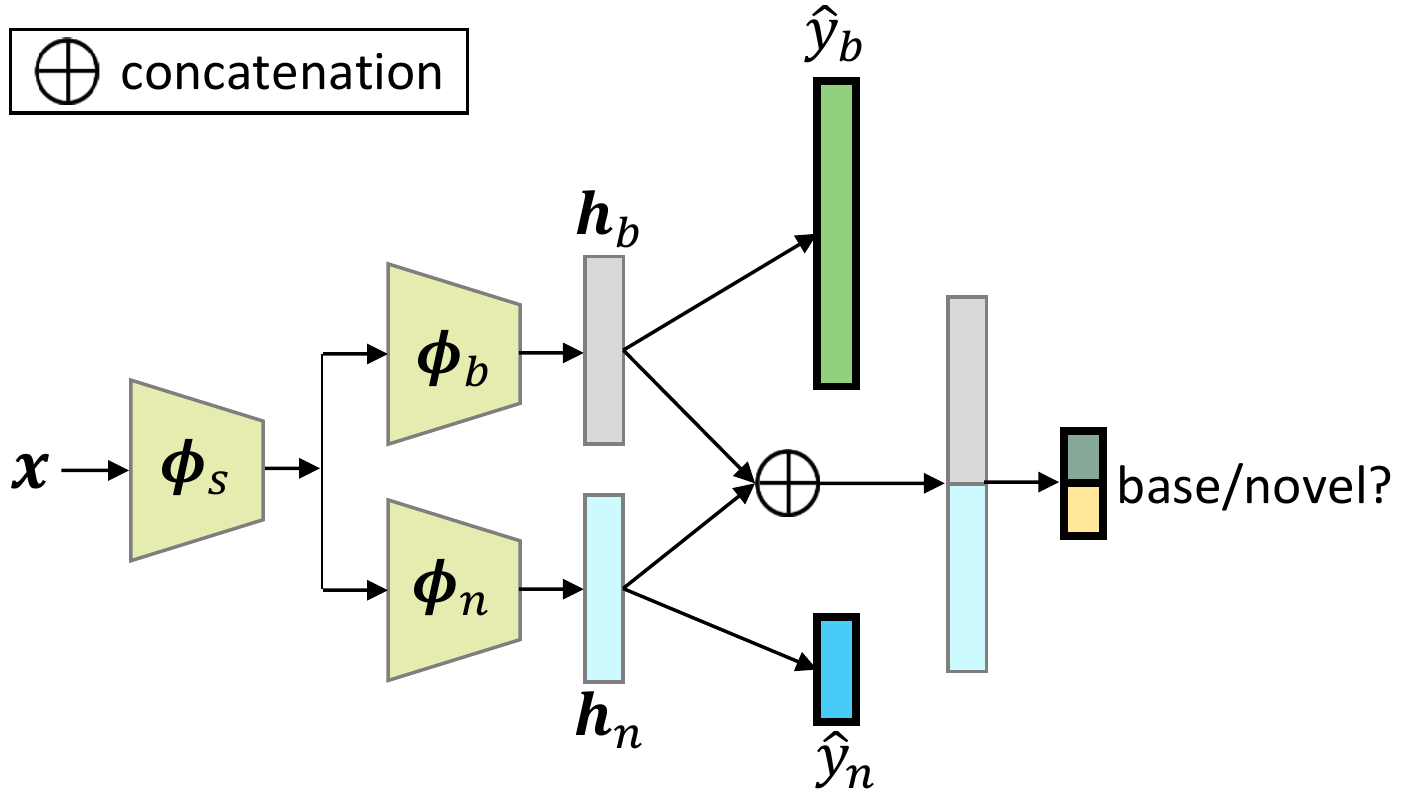}
        \subcaption{\footnotesize Learning-based routing baseline.}
        \label{fig:route_learn}
    \end{minipage}
    \begin{minipage}[h]{0.5\linewidth}
        \centering
      \includegraphics[width=\linewidth]{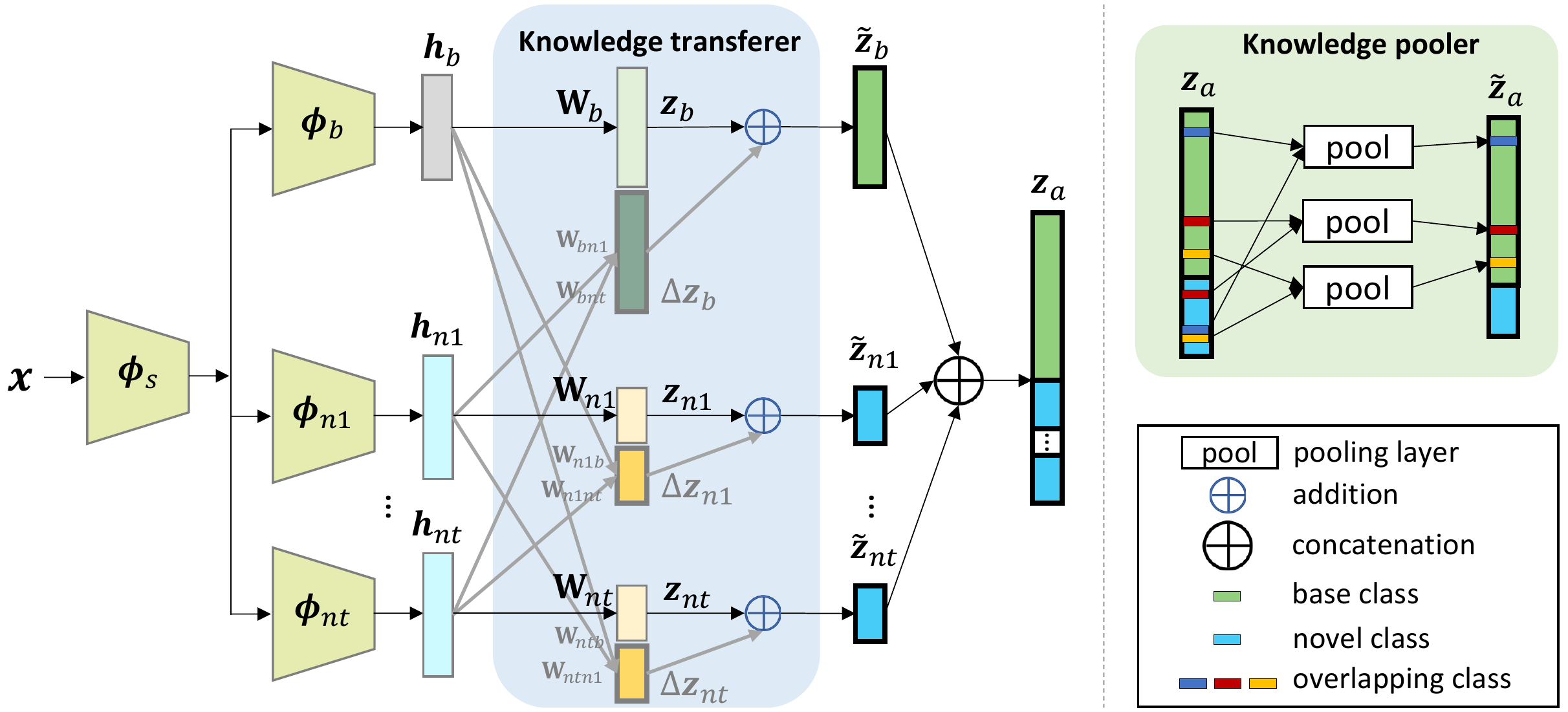}
        \subcaption{\footnotesize General score fusion network.}
        \label{fig:model}
    \end{minipage}
    \end{minipage}
    \caption{{\bf Fusion methods.} (a-b) Routing as a baseline. (c) Our proposed network for score fusion.}\label{fig:fuse}
\end{figure*}


\section{Fusion: unifying base and novel classifiers}\label{sec:fusion}
\subsection{Baseline using routing}
We analyze fusion baselines with only one incremental step, although they can naturally extend to additional steps. We first explore two intuitive baselines that 
rely on each classifier's prediction and a 
routing function $\hat{r}(\cdot)\in \{0,1\}$ to decide whether a sample is from $\mathcal{D}_b$ or $\mathcal{D}_n$. The final prediction $\hat{y}$ is then assigned by $\hat{y}_{b}$ if $\hat{r}(\mathbf{x})=0$ and $\hat{y}_{n}$ if $\hat{r}(\mathbf{x})=1$,
where $\hat{y}_d = \argmax_l \hat{p}^{(l)}(\mathbf{x};\mathbf{\Phi}_s,\mathbf{\Phi}_d,\mathbf{W}_d)$, for $d\in \{b,n\}$. We explore two routing functions:

\point{Confidence-based routing} (Fig.~\ref{fig:route_conv}) uses the confidence score of individual classifiers as the proxy where the routing function is defined as $\hat{r}(\mathbf{x})=\mathbbm{1}_{[Conf_{b}<Conf_{n}]}$, where $\mathbbm{1}_{[\cdot]}$ is the indicator function, and the confidence score $Conf_{d}=\max_l \hat{p}^{(l)}(\mathbf{x};\mathbf{\Phi}_s,\mathbf{\Phi}_d,\mathbf{W}_d)$, for $d\in \{b,n\}$. 

\point{Learning-based routing} (Fig.~\ref{fig:route_learn}) directly learns a 
routing classifier with available data. Following prior work in CIL~\cite{icarl}, we keep few-shot exemplars $\mathcal{E}$ from all past incremental steps in the memory to supplement the novel data $\mathcal{D}_n$. The routing classifier is formulated as
\begin{align}
     &\hat{r}(\mathbf{x}) = \sigma(\mathbf{W}_r^T(\mathbf{h}_b\oplus \mathbf{h}_n))~,
\end{align}
where $\oplus$ denotes vector concatenation. For $d\in \{b,n\}$, the feature from each feature extractor branch, $\mathbf{h}_d=h(\mathbf{x};\mathbf{\Phi}_s,\mathbf{\Phi}_d)$, is frozen to preserve the information, and the linear weights $\mathbf{W}_r\in\mathbb{R}^{2k\times2}$ are the only parameters for learning the routing classifier.

The routing loss is a binary cross-entropy loss, \ie 
\begin{align}
    &l_{rt}(\mathbf{x}, r) = -(1-r) \log \hat{r}^{(0)}(\mathbf{x})-r \log \hat{r}^{(1)}(\mathbf{x})~,
\end{align}
where $r=\mathbbm{1}_{[\mathbf{x}\in\mathcal{D}_n]}$ is the split label for $\mathbf{x}$.  
However, there is a large base-novel sample imbalance due to $|\mathcal{D}_n|\gg |\mathcal{E}|$. 
To address this, we re-balance the class losses by re-weighting:
\begin{align}
    &L_{rt-bal} = \frac{1}{2|\mathcal{E}|}\sum_{\mathbf{x}_i \in \mathcal{E}} l_{rt}(\mathbf{x}_i, r_i)+\frac{1}{2|\mathcal{D}_n|}\sum_{\mathbf{x}_i \in \mathcal{D}_n} l_{rt}(\mathbf{x}_i, r_i).\label{eq:rt-bal}
\end{align}

While routing is a way to get the prediction for all classes in $\mathcal{Y}_a$, these baselines cannot produce a unified probability distribution for all classes. Also, the routing function's prediction error will propagate to the final class prediction. 

\subsection{General score fusion network}
Our proposed score fusion network (Figure~\ref{fig:model}) integrates the knowledge and output of all branches and generates a unified probability distribution. Modifying the existing well-trained classifiers with few samples leads to overfitting while being susceptible to class imbalance in the limited data, but freezing everything rules out knowledge transfer opportunities between branches. In addition, predictions regarding overlapping classes need unified scores from base and novel classifiers. We introduce knowledge-preserving transfer, overlap knowledge integration schemes, and balanced optimization to address these issues. 

\point{Knowledge-preserving transfer.}  After the stage-I training, we obtain expert models for each step $\{\mathbf{\Phi}_d, \mathbf{W}_d\}$ (for $d\in \{b,n1,\dots,nt\}$), where the probability of the individual classifier is computed by applying the softmax function on the logit score $\mathbf{z}_{d} = \mathbf{W}_d^T \mathbf{h}_d$, $\mathbf{W}_d\in\mathbb{R}^{k\times|\mathcal{Y}_d|}$.
To prevent overfitting to the small dataset of $\mathcal{E} \cup \mathcal{D}_{nt}$, 
we propose to also freeze all classifier weights (\ie $\mathbf{W}_b,\mathbf{W}_{n\tau}$, in addition to $\mathbf{\Phi}_s,\mathbf{\Phi}_b,\mathbf{\Phi}_{n\tau}$, $\tau \in [1,t]$) to preserve their capability of distinguishing classes within the same split. To additionally enable the knowledge transfer between the two splits, we use $\mathcal{E} \cup \mathcal{D}_{nt}$ to learn
$\mathbf{W}_{dd'}\in\mathbb{R}^{k\times|\mathcal{Y}_{d}|}$, $d,d' \in \{b,n1,\dots,nt\}$, $d \neq d'$ (randomly initialized), connecting $d'$ branch's features to $d$ branch's logits, which are used to learn the delta logits for knowledge transfer, to be added to the final logits for the $d$ branch: 
\begin{align}
    \triangle \mathbf{z}_{d} &= \sum_{d' = b, n1, \dots, nt}^{d' \neq d} \mathbf{W}_{dd'}^T \mathbf{h}_{d'}, \\
    \tilde{\mathbf{z}}_d &= \mathbf{z}_d + \triangle \mathbf{z}_{d}.
\end{align}

\point{Overlap knowledge integration.} To get a unified classifier with probability distribution for all the classes, we can combine the logit scores of the base and novel branches by concatenation, \ie
$\mathbf{z}_a = \tilde{\mathbf{z}}_b \oplus \tilde{\mathbf{z}}_{n1} \oplus \dots \oplus \tilde{\mathbf{z}}_{nt}$ where $\tilde{\mathbf{z}}_d\in \mathbb{R}^{|\mathcal{Y}_d|}$, $d \in \{b,n1,\dots,nt\}$, and obtain the posterior probability with $\sigma(\mathbf{z}_a)$. However, when overlapping classes exist ($\mathcal{Y}_{d} \cap \mathcal{Y}_{d'} \ne \emptyset, \exists d, d'$), they appear in $\tilde{\mathbf{z}}_a$ multiple times.
We apply a knowledge pooler to $\mathbf{z}_a\in \mathbb{R}^{\sum_d|\mathcal{Y}_d|}$ to get the final logit $\tilde{\mathbf{z}}_a\in \mathbb{R}^{|\mathcal{Y}_a|}$, as illustrated in Figure~\ref{fig:model} (right), that either max-pools or average-pools the multiple logit scores for each overlapping class. 
Note that $\tilde{\mathbf{z}}_a=\mathbf{z}_a$ when $\mathcal{Y}_{d} \cap \mathcal{Y}_{d'}=\emptyset, \forall d, d'$. From our experiments, max pooling performs better than average pooling, since the branches do not always simultaneously output high responses to a sample on its class, especially when the data of the same class are very different in base and novel splits.  

\point{Balanced optimization.} With the final logit score $\tilde{\mathbf{z}}_{a,i}$ for each sample $\mathbf{x}_i\in \mathcal{E}\cup \mathcal{D}_{nt}$, the parameters $\mathbf{W}_{dd'}$ can be optimized with the loss of eq.~(\ref{eq:ce}). However, since $|\mathcal{D}_{nt}|\gg |\mathcal{E}|$, the training will be dominated by the novel classes. To balance the probability estimation, we sample a subset $\mathcal{B}\subset \mathcal{E}\cup \mathcal{D}_{nt}$ uniformly over all the classes, where each class in $\mathcal{B}$ has equal number of samples. With this class-balanced sampling, the classification loss becomes
\begin{align}
L_{cls} = \frac{1}{|\mathcal{B}|}\sum_{i=1}^{|\mathcal{B}|} -\log \hat{p}^{(y_i)}(\mathbf{x}_i)\quad \hat{p}(\mathbf{x})=\sigma(\tilde{\mathbf{z}}_{a}).\label{eq:cls}
\end{align}
However, in addition to the sample imbalance, our number of classes is also highly imbalanced. Since $|\mathcal{Y}_b|\gg |\mathcal{Y}_{n\tau}|$, base logits have more chances of being the largest than novel logits, so eq.~(\ref{eq:cls}) will favor one of the base classes, which may or may not be desired depending on the application. To control base and novel logit balance, we explore two regularization mechanisms.

First, we explicitly train to balance the largest base logit and the largest novel logit scores, by adding a routing auxiliary loss over the maximum score from each split. The routing classifier can be defined as
\begin{align}
    \hat{r}(\mathbf{x}) = \sigma(\mathbf{W}_{r,aux}^T(\max_l \tilde{\mathbf{z}}_b^{(l)}\oplus \max_l\tilde{\mathbf{z}}_{n1}^{(l)}\oplus\dots\oplus\max_l\tilde{\mathbf{z}}_{nt}^{(l)}))~,\label{eq:fusion_routing}
\end{align}
where $\mathbf{W}_{r,aux}\in\mathbb{R}^{(t+1)\times(t+1)}$ is the linear routing classifier's weights. The full loss then becomes 
\begin{align}
    L_{total} = (1-\alpha) \cdot L_{cls}+\alpha \cdot L_{rt-bal}~, \label{eq:final_loss}
\end{align}
with $L_{cls}$ from eq.~(\ref{eq:cls}), $L_{rt-bal}$ from eq.~(\ref{eq:rt-bal}) using $\hat{r}$ from eq.~(\ref{eq:fusion_routing}), and $\alpha$ is a loss weight hyperparameter.  


Second, during training we normalize and scale $\mathbf{h}_{d'}$ for base samples by a factor of $\beta\in[0,1]$ before feeding it into $\mathbf{W}_{bd'}$, $d'\neq b$. This prevents the training from drastically influencing base classes, but limits the knowledge transfer from novel branch features to base classes.

Like all imbalanced learning problems, there is a trade-off between the performance of base and novel classes. Using our two regularization mechanisms provides the flexibility to optimize for the metric that matters more for each customized application. The behavior of these mechanisms will be discussed further in the experiment section.

\section{Experiments}
In this section, we compare our results with baselines and state-of-the-art CIL methods. We also show ablation studies on the effect of different components and analyze our results in different practical settings. Code is available\footnote{\url{https://github.com/amazon-research/sp-cil}}.
\subsection{Experimental setup}\label{sec:setup}
\point{Dataset.} To study the behavior of strong pre-trained models, we create several data splits from ImageNet~\cite{imagenet} to simulate different practical scenarios. For {\it disjoint-CIL} with disjoint base and novel classes, we first perform our main analysis with one incremental step. Unless otherwise noted, we randomly select 800 base classes and 40 novel classes. Variations are tested in ablation studies. We then test on an existing 10-step asymmetric split scheme with 500 base classes and ten steps of 50 classes each~\cite{podnet}, originally designed to test short but numerous incremental steps. 

For \textit{overlapping-CIL}, where some or all novel classes overlap with base classes, we experiment with three different base-novel class splits (one incremental step). {\bf 1) random class split}: 800 base classes, 40 novel classes with 5 overlapping classes among them. {\bf 2) domain-changing split}: ImageNet classes are grouped into two categories, animate and inanimate. Inanimate and animate classes are taken as base and novel classes respectively, with 5 overlapping classes that are randomly selected. {\bf 3) style-changing split}: To simulate this scenario with currently available annotations of ImageNet, we select five pairs of classes that are semantically similar to each other, merge each pair into a single class, and use them as five overlapping classes. For example, ``pembroke" and ``cardigan" can be merged into ``corgi", and ``electric guitar" and ``acoustic guitar" can be merged into ``guitar". Non-overlapping novel and base classes are chosen randomly.

For overlapping classes in (1) and (2), samples need to be distributed between base and novel splits. We explore splitting each class either \textbf{randomly} or by \textbf{unsupervised clustering} using K-Means ($K=2$) on features extracted from a full-ImageNet trained ResNet10 penultimate layer. After splitting all 1000 classes into two splits, we use one split for base classes and the other split for novel classes. That is, the 800 base classes are trained on roughly half of the 800 classes' data.

\point{Metrics.} In prior work, the number of base and novel classes are usually balanced (\eg $|\mathcal{Y}_b|=|\mathcal{Y}_n|=50$), so it is natural to simply evaluate using the accuracy over all the classes, \ie $Acc_{all}=\mathcal{A}(\mathcal{D}_{test})$, where $\mathcal{D}_{test}$ is the testing set and $\mathcal{A}(S)=\frac{1}{|S|}\sum_{(x,y)\in\mathcal{S}}\mathbbm{1}_{[y=\argmax_k \hat{p}^{(k)}(\mathbf{x})]}$ either at $t=T$ or averaged over all incremental steps (identical to incremental accuracy). We follow the version implemented in~\cite{podnet} that includes the base step $t=0$ to facilitate comparison. However, in our setting of strong pre-trained models, $|\mathcal{Y}_b|\gg|\mathcal{Y}_n|$, \eg 800 and 40. In this case, the overall accuracy will be dominated by the base class performance. A model with high overall accuracy is not guaranteed to perform well for the novel classes. Hence, in addition to the overall accuracy, we also present the accuracy in each split. When there is overlap, we use $Acc_{base}=\mathcal{A}(\{(x,y):(x,y)\in\mathcal{D}_{test},y\in\mathcal{Y}_b\setminus{(\mathcal{Y}_b\cap \mathcal{Y}_n)}\})$, $Acc_{novel}=\mathcal{A}(\{(x,y):(x,y)\in\mathcal{D}_{test},y\in\mathcal{Y}_n\setminus{(\mathcal{Y}_b\cap \mathcal{Y}_n)}\})$ and $Acc_{ovlp}=\mathcal{A}(\{(x,y):(x,y)\in\mathcal{D}_{test},y\in{\mathcal{Y}_b\cap \mathcal{Y}_n}\})$. Note that the classification is still predicted among all classes. An aggregate metric to balance among the per split accuracy is the average of them: $Acc_{avg}=\frac{\sum_{d\in\{b, n1, \dots, nt\}} Acc_{d}}{t+1}$ either at the final $t=T$ step or follow the 500+50x10 split implementation~\cite{podnet} by averaging over all steps ($Acc_{avg}^{t=0..10}$).
When there is class overlap, we use $Acc_{avg}=(Acc_{base}+Acc_{novel}+Acc_{ovlp})/3$ for partial overlap and $Acc_{avg}=(Acc_{base}+Acc_{ovlp})/2$ for full overlap (${Y}_n \subseteq {Y}_b $).


\point{Learning.} For stage-II training, 10 exemplars per class are randomly selected to create a class-balanced split, and the experiments are repeated with different random seeds. Hyperparameters $\alpha$ and $\beta$ are selected with a grid search conducted with the validation set as per the metric of interest. For fairness, all compared methods use the same memory constraint and dataset sizes. Dataset subsampling are also identical except where results are from other papers.

\subsection{Disjoint-CIL}
We first study the most common scenario in the literature, disjoint-CIL, where the set of base and novel classes are disjoint. In the following, we present the comparison to the baselines, along with extensive ablations and analysis of the proposed method.  

\begin{figure}
    \centering

    \includegraphics[width=0.49\linewidth]{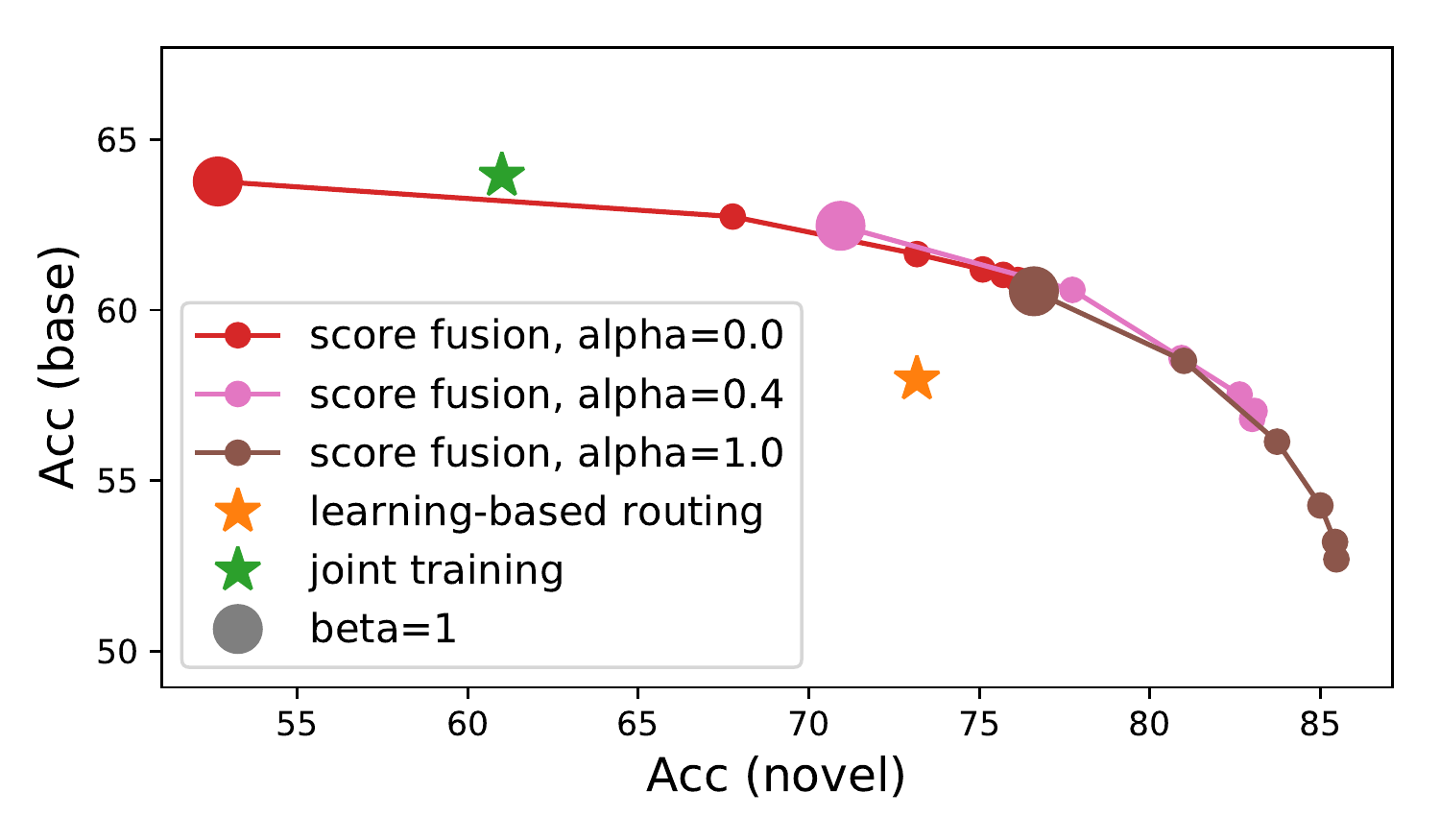}
    \caption{\textbf{Design space.} Score fusion performs comparably to joint training and outperforms learning based routing. Optimal $\alpha$ and $\beta$ values depend on the desired balance between base and novel performance.}
    \label{fig:design}
\end{figure}

\point{Design choices.} We first study the balancing behavior of the two regularization methods in score fusion, routing loss and feature scaling. For demonstration, we show a hyperparameter sweep in Figure~\ref{fig:design} with $\alpha \in \{0,0.4,1\}, \beta \in \{0,0.2,0.4,0.6,0.8,1\}$. Actual hyperparameters are selected on validation data instead. As we increase $\alpha$ (routing loss weight) and decrease $\beta$ (scaling factor), our novel class performance is boosted significantly, while the drop in base class accuracy is comparably minor. The effects of $\alpha$ and $\beta$ are similar, but changing both gives us the flexibility to reach the full range of base-novel balance operating points.  

The optimal hyperparameter selection depends on the specific design choices and desired metric, such as balancing between base and novel (best $Acc_{avg}$) or prioritizing novel classes (best $Acc_{novel}$). In the following experiments, 
we will present the score fusion results optimizing for these practical scenarios: {\it best-$Acc_{all}$}, {\it best-$Acc_{avg}$} and {\it best-balanced} (optimize $\frac{Acc_{all}+Acc_{avg}}{2}$).

\begin{table*}[t!]
    \centering
    \caption{Comparison to SOTA class-incremental learning methods. Our method outperforms without additional parameters and pushes the performance further with additional parameters.}\label{tab:sota}
    \setlength{\tabcolsep}{7pt}
    \resizebox{1\linewidth}{!}{
    \begin{tabular}{c|ccccc|ccccc}
    \toprule
     & \multicolumn{5}{c|}{ResNet10} & \multicolumn{5}{c}{ResNet18} \\
     Method & \# of params & $Acc_{all}$ & $Acc_{base}$ & $Acc_{novel}$ & $Acc_{avg}$ 
        & \# of params & $Acc_{all}$ & $Acc_{base}$ & $Acc_{novel}$ & $Acc_{avg}$ \\ \midrule
     fine-tuning                      & 4.9M  & 4.18  & 0.01  & 87.63 & 43.82
        & 11.2M & 4.25  & 0.00  & 89.37 & 44.68 \\
     LwF~\cite{lwf}                  &       & 9.50  & 5.54  & 88.53 & 47.04
        &       & 9.50  & 5.46  & 90.30 & 47.88 \\
     iCaRL~\cite{icarl}              &       & 16.26 & 13.91 & 63.40 & 38.66 & & 10.65 & 8.15 & 60.80 & 34.78 \\
     BiC~\cite{bic}                  &       & 30.30 & 27.55 & 85.20 & 56.38
        &       & 31.50 & 28.75 & 86.60 & 57.68 \\
     WA~\cite{wa}                    &       & 51.33 & 52.33 & 31.40 & 41.87
        &       & 54.79 & 55.17 & 47.20 & 51.19 \\
     DER w/o P~\cite{der}            & 9.8M  & 52.31 & 52.43 & 50.10 & 51.27 
        & -     & -     & -     & -     & -     \\\midrule
     score fusion (ours) best-$Acc_{all}$ & 8.6M  & \textbf{63.24} & 63.77& 52.67 & 58.22
        & 19.6M & \textbf{69.45} & 70.01 & 58.13 & 64.07 \\
     score fusion (ours) best-balanced    &       & 62.15 & 61.49 & 75.37 & 68.43
        &       & 67.36	& 66.61	& 82.37	& 74.49 \\
     score fusion (ours) best-$Acc_{avg}$ &       & 58.90 & 57.73 & 82.40 & \textbf{70.06} 
        &       & 65.83	& 64.85	& 82.50	& \textbf{75.17} \\\midrule
score fusion (ours, fc-only) best-$Acc_{all}$      & 4.9M & 62.65 & 63.56 & 44.53 & 54.05 & 11.2M & 68.79 & 69.58 & 53.07 & 61.32 \\
score fusion (ours, fc-only) best-balanced         &      & 61.01 & 60.81 & 65.07 & 62.94 &       & 66.76 & 66.50 & 71.83 & 69.17 \\
score fusion (ours, fc-only) best-$Acc_{avg}$      &      & 57.91 & 57.24 & 71.57 & 64.40 &       & 65.89 & 65.49 & 73.77 & 69.63 \\ \midrule
     joint learning (oracle)                  & 4.9M  & 63.80 & 63.94 & 61.00 & 62.47
        & 11.2M & 70.32 & 70.43	& 68.20	& 69.32 \\\bottomrule
    \end{tabular}
    }
\end{table*}

\begin{table}[t!]
    \centering
    \caption{Multiple novel steps disjoint-CIL with ResNet10, 500 base classes with ten 50-class novel increments, random class split. For ResNet18, BIC and PODNet results are from~\cite{podnet}, and we use their experimental setup and 20/class memory constraints.
    }
    \setlength{\tabcolsep}{5pt}
    \resizebox{1\linewidth}{!}{
    \begin{tabular}{c|cc|cc}
    \toprule
    Network & \multicolumn{2}{c|}{ResNet10} & \multicolumn{2}{c}{ResNet18} \\
    Method & inc. acc. & $Acc_{avg}^{t=0..10}$ & inc. acc. & $Acc_{avg}^{t=0..10}$ \\ \midrule
    BiC~\cite{bic}                                 & -- & -- & 44.31 & -- \\
    PODNet~\cite{podnet}                              & -- & -- & 64.13 & -- \\\midrule
score fusion (ours) best-$Acc_{all}$             & \textbf{62.70} & 61.45 & \textbf{67.48} & 65.66 \\
score fusion (ours) best-balanced                & 61.03 & 65.34 & 65.95 & 69.40 \\
score fusion (ours) best-$Acc_{avg}$             & 55.31 & \textbf{66.23} & 61.06 & \textbf{70.67}\\ \bottomrule
    \end{tabular}
    }
    \label{tab:multistep}
\end{table}

\point{Comparisons to CIL methods.} In Table~\ref{tab:sota} (one novel step) and Table~\ref{tab:multistep} (multiple novel steps), we compare to state-of-the-art class-incremental learning methods and the joint training oracle that uses all base data but unavailable to us. For score fusion, we report the results of three operating points, the one with the best $Acc_{all}$, the best $Acc_{avg}$, and the most balanced performance of these two metrics. Fine-tuning and LwF~\cite{lwf} are simple baselines that do not adopt image replay and perform especially badly. iCaRL~\cite{icarl}, BiC~\cite{bic}, PODNet~\cite{podnet}, and WA~\cite{wa} keep few-shot examplars to mitigate forgetting,
but these methods still underperform on base classes since they use a small amount of data to modify the well-trained representation layers. We outperform other methods under the multiple novel step benchmark as well, including PODNet~\cite{podnet} which proposed the benchmark. DER w/o P~\cite{der} is a recent approach that freezes the feature extractor of the base network and creates an identical branch to learn novel features. While similar to the proposed method, the architecture of the additional feature extractor branch and the training objectives are different. In addition, they relearn the final linear classifier with limited data, which we show is less effective in our ablation studies (Table~\ref{tab:disjoint}). 
Compared with existing CIL methods with the same backbone, the proposed method achieves significantly better performance in both $Acc_{all}$ and $Acc_{avg}$ and generalize well into multiple novel step scenarios. 

For fair comparison, we also present our results branching at the classifier layer (``fc-only'') to bring our number of network parameters on par with most compared methods, \ie $\mathbf{\Phi}_s=\mathbf{\Phi}$, $\mathbf{\Phi}_b=\mathbf{\Phi}_n=\emptyset$, and $\mathbf{h}_b=\mathbf{h}_n$. Our results still outperform all compared methods, and interestingly, our ResNet10 results outperform others' ResNet18 results despite using fewer parameters.

\begin{table}[t!]
    \centering
    \caption{Disjoint-CIL analysis with ResNet10, 40 novel classes using random class split. (Results for ResNet18/ResNet50 architectures and for 200 novel classes are in appendix)
    \ginanote{split2, 800-40-0, resnet10}
    }
    \setlength{\tabcolsep}{5pt}
    \resizebox{1\linewidth}{!}{
    \begin{tabular}{c|c|cc|c}
    \toprule
    Method & $Acc_{all}$ & $Acc_{base}$ & $Acc_{novel}$ & $Acc_{avg}$ \\ \midrule
    confidence-based routing                    & 41.58 & 39.26 & 88.00 & 63.63 \\
    learning-based routing w/ $L_{rt-bal}$   & 58.69 & 57.97 & 73.17 & 65.57 \\
    oracle routing w/ $L_{rt-bal}$           & 58.57 & 57.48	& 80.50 & 68.99 \\ \hline
    FeatCat+RT                                  & 55.94 & 56.08 & 53.27 & 54.67 \\
    LogitCat+RT                                 & 58.36	& 58.56	& 54.30	& 56.43 \\ 
    LogitCat+FT                                 & 59.39	& 59.34	& 60.57	& 59.95 \\ \hline
    FA (ours) + BiC~\cite{bic}                           & 63.21	& 63.66 & 54.20 & 58.93 \\
    FA (ours) + WA~\cite{wa}                             & 62.77	& 64.19	& 34.30	& 49.25 \\ \midrule
score fusion (ours) best-$Acc_{all}$             & \textbf{63.24} & 63.77 & 52.67 & 58.22 \\
score fusion (ours) best-balanced                & 62.15 & 61.49 & 75.37 & 68.43 \\
score fusion (ours) best-$Acc_{avg}$             & 58.90 & 57.73 & 82.40 & \textbf{70.06} \\ \midrule
    joint learning (oracle)                              & 63.80	& 63.94	& 61.00	& 62.47 \\\bottomrule
    \end{tabular}\label{tab:disjoint}
    }
\end{table}

\point{Comparisons to fusion baselines.} In Table~\ref{tab:disjoint}, we explore two lines of fusion methods in stage-II training, routing, and score fusion. We show that our observations hold for different backbones (ResNet18/ResNet50) and more novel classes in appendix.

Among routing methods, confidence-based routing has the worst $Acc_{base}$, and around half of the base samples are misclassified as novel. Learning-based routing improves the result for $Acc_{base}$, and reaches a more balanced performance. The oracle routing performance (trained on all base and novel data rather than 10-shot) is also shown in the table for reference, and it mainly improves in novel classes and the gap in $Acc_{base}$ remains.  

We then compare the score fusion method with a set of baselines in the proposed framework in Table~\ref{tab:disjoint}.  {\it FeatCat + RT} retrains a linear classifier with features $\mathbf{h}_b$ and $\mathbf{h}_n$ concatenated. {\it LogitCat} concatenates logits $\mathbf{z}_{base}$ and $\mathbf{z}_{novel}$, and retrains ({\it RT}) or fine-tunes ({\it FT}) the linear weights $\mathbf{W}_b$ and $\mathbf{W}_n$. Since the original linear weights were already trained using a large amount of data, retraining them from scratch on 10-shot with feature or logit concatenation leads to inferior results than fine-tuning the linear weights, as shown in the table. This supports our idea of preserving the weights of the original linear classifiers. 

Next, we apply two score fusion baselines, BiC~\cite{bic} and WA~\cite{wa}, to our network (FA) after stage-I training. Both methods estimate a scalar to balance between the base and novel logits, and they estimate it by validation data and weight norms respectively. Using these methods with our network preserves the base class performance better than {\it RT} or {\it FT} methods, and FA+BiC on ResNet10 even performs similarly to ours (but not on ResNet18). Although we note that using BiC and WA directly performs poorly as shown in Table~\ref{tab:sota}. 


\point{Robustness.} To test the robustness of the proposed method, we study score fusion with a larger number of novel classes and deeper networks. Results of 800 base and 200 novel classes instead of 40 and the results for different architectures are in appendix. The same trends in Table~\ref{tab:disjoint} hold.

\begin{table}[t!]
    \centering
    \caption{Disjoint-CIL with ResNet10, 40 novel classes using the inanimate-animate split (wider base-novel gap). Score fusion can generalize to fine-tuning more/less layers in stage-I if necessary. }
    \label{tab:layer3-4}
    \setlength{\tabcolsep}{5pt}
    \resizebox{1\linewidth}{!}{
    \begin{tabular}{c|c|c|cc|c}
    \toprule
  Trainable & Method & $Acc_{all}$ & $Acc_{base}$ & $Acc_{novel}$ & $Acc_{avg}$ \\ \midrule
layer3,4 & score fusion (ours) best-balanced                & 57.08 & 55.27 & 83.10 & 69.18 \\
layer4   & score fusion (ours) best-balanced                & 57.12 & 55.50 & 80.37 & 67.93 \\
layer4 (conv2 only) & score fusion (ours) best-balanced                & 57.15 & 55.73 & 77.37 & 66.55 \\
        \bottomrule

    \end{tabular}
    }
\end{table}


\point{Generalization.} When the base and novel data are very different, \eg in Table~\ref{tab:layer3-4} where we use the domain-changing split (inanimate as base classes and animate for novel), fine-tuning more layers in the first stage may produce higher accuracy on the novel data. We show that our score fusion generalizes to different ways of branching the network.

\subsection{Overlapping-CIL}    
In practice, base and novel classes may not be mutually exclusive. In this section, we study the overlapping-CIL scenario where a subset of base and novel classes overlap. As a recap of Section~\ref{sec:setup}, we analyze five different ways to split base and novel classes and samples within each class.
Unless otherwise noted, we test on ResNet10 with 800 base, 40 novel with 5 overlapping classes. 

Table~\ref{tab:style} shows the result when the style of the overlapping classes changes from base to novel. We simulate this by merging pairs of similar classes that are siblings in the WordNet hierarchy~\cite{wordnet,Wu20DeepRTC}, and placing the two classes in each pair into the base and novel split respectively. Compared to average pooling, applying max pooling to the logits of overlapping classes performs better in $Acc_{ovlp}$ with little drawback in $Acc_{base}$. Hence, in our score fusion results, we adopt max pooling to fuse the logits. Confidence-based routing performs the worst for the base classes. Our score fusion achieves the best performance for both splits even when the samples of overlapping classes are in a different style from base to novel. 

Due to space constraints, we show in the appendix results for the random base-novel-overlap split scenario and the domain change (inanimate and animate, respectively, and 5 inanimate classes as overlapping classes) scenario. The conclusions are identical and our score fusion generalizes to both scenarios, with overlapping class samples either randomly split or clustered.



\begin{table}[t!]
    \centering
    \caption{Overlapping-CIL results of style-changing splits (merge similar classes into one). See appendix for overlapping-CIL with random and domain-changing split. \ginanote{split8, 800-40-5, Resnet10}}\label{tab:style}
    \setlength{\tabcolsep}{3pt}
    \resizebox{1\linewidth}{!}{
    \begin{tabular}{c|c|ccc|c}
    \toprule
    Method & $Acc_{all}$ & $Acc_{base}$ & $Acc_{novel}$ & $Acc_{ovlp}$ & $Acc_{avg}$ \\ \midrule
    confidence-based routing                    & 40.70 & 38.17 & 86.97 & 80.00 & 68.38 \\
    learning-based routing w/ $L_{rt-bal}$   & 59.14	& 58.63	& 72.91 & 51.73	& 61.09 \\
    oracle routing w/ $L_{rt-bal}$           & 59.57 & 58.83 & 78.63 & 51.20	& 62.89 \\ \hline
    logit concatenation (avg pool)              & 63.39	& 63.63 & 64.57 & 40.40 & 56.20 \\
    logit concatenation (max pool)              & 63.50 & 63.38 & 64.46 & 69.60 & 65.81 \\ \midrule
score fusion (ours) best-$Acc_{all}$             & \textbf{63.81} & 64.35 & 53.76 & 56.13 & 58.08 \\
score fusion (ours) best-balanced                & 61.02 & 60.02 & 79.58 & 75.60 & 71.73 \\
score fusion (ours) best-$Acc_{avg}$             & 57.58 & 56.18 & 83.77 & 77.73 & \textbf{72.56} \\ \midrule
    joint learning (oracle)                              & 64.33 & 64.28 & 62.06 & 76.80 & 67.71 \\\bottomrule
    \end{tabular}}
\end{table}


Additionally, Table~\ref{tab:full_overlap} shows an extreme full-overlap scenario where all novel classes are overlapping with base classes (random base-novel split, split overlapping classes by clustering). This simulates the practical scenario where one needs to adapt a portion of existing classes to a new domain. We compare to prior incremental learning work. These methods underperform due to changing the backbone network weights. Even Rainbow Memory~\cite{rainbowmem} which explicitly tackles fully overlapping classes, underperforms, partially because it assumes the distribution of each overlapping class does not change. In comparison, our method generalizes well to this scenario.

\begin{table}[t!]
    \centering
    \caption{Fully-overlapping-CIL with all 40 novel classes overlapping with the 800 base classes, \ie $\mathcal{Y}_n\subset \mathcal{Y}_b$.\ginanote{split9, 800-40-40, resnet10}}
    \setlength{\tabcolsep}{5pt}
    \resizebox{1\linewidth}{!}{
    \begin{tabular}{c|c|c|cc|c}
    \toprule
    Backbone & Method & $Acc_{all}$ & $Acc_{base}$ & $Acc_{ovlp}$ & $Acc_{avg}$ \\ \midrule
    ResNet10 & pre-trained model                       & 57.69 & 58.03 & 51.20 & 54.62 \\ \cline{2-6}
        & confidence-based routing                    & 35.43 & 33.01 & 81.50 & 57.25 \\
        & learning-based routing w/ $L_{rt-bal}$      & 52.82 & 52.12 & 66.10 & 59.11 \\ \cline{2-6}
        & fine-tuning                                  & 3.86 & 0.00 & 77.20 & 38.6 \\
        & iCaRL~\cite{icarl}           & 20.23 & 19.77 & 28.90 & 24.34 \\
        & RM~\cite{rainbowmem}         & 21.10 & 18.12 & 77.70 & 47.91 \\ \cline{2-6}
& score fusion (ours) best-$Acc_{all}$             & \textbf{57.82} & 57.80 & 58.23 & 58.02 \\
& score fusion (ours) best-balanced                & 55.66 & 54.81 & 71.90 & 63.36 \\
& score fusion (ours) best-$Acc_{avg}$             & 53.42 & 52.28 & 75.13 & \textbf{63.70} \\ \cline{2-6}
        & joint learning (oracle)                              & 47.63 & 47.32 & 53.60 & 50.46 \\\bottomrule
    ResNet18 & iCaRL~\cite{icarl}     & 18.12 & 17.04 & 38.60 & 27.82 \\
        & RM~\cite{rainbowmem}         & 21.89 & 28.78 & 80.90 & 49.84 \\ \midrule
    \end{tabular}\label{tab:full_overlap}
    }
\end{table}

\section{Conclusion and Limitations}
In this work, we investigated CIL in the context of a pre-trained model with large number of base classes. We showed how branching can be an effective solution for learning novel data when using a strong pre-trained model and how it can preserve the learning with the old data. Furthermore, we discuss a novel score fusion method that uses both feature and classifier information from old and novel networks and generates a unified classifier. This approach leads to state-of-the-art results for CIL with large number of base classes. Our method can further be improved by using distillation approach to reduce memory footprint. 

{\small
\bibliographystyle{ieee_fullname}
\bibliography{egbib}
}

\appendix

\section{Additional experiments}

\subsection{Disjoint-CIL}

\point{Robustness.} To test the robustness of the proposed method for different backbones, we study score fusion with ResNet18 and ResNet50. The results are shown in Tables~\ref{suptab:deepnet18}, and \ref{suptab:deepnet50}. As shown for the ResNet10 backbone in the main paper, the score fusion approach results in better performance even for deeper backbones, but the relative trends remain the same. In addition, We also study score fusion with a larger number of novel classes. In the main paper, we showed results for 800 base and 40 novel classes. Here, we show the results for 800 base and 200 novel classes in Table~\ref{suptab:disjoint200}. The results for 200 classes are similar to the ones for 40 classes (refer Table~\ref{tab:disjoint}). This shows that our score fusion method can be generalized to larger number of novel classes.

\subsection{Overlapping-CIL}    
As a recap of Section~\ref{sec:setup}, in the overlapping-CIL scenario where a subset of base and novel classes overlap, we analyze three different ways to split base and novel classes and two different ways to split the samples within each class. In particular, we showed the results for splits based on style in Table~\ref{tab:style}. In this section, we show the results for the rest of the splits. 

We show in Table~\ref{suptab:randcls} results for random split and within that splitting the samples either randomly or using clustering. In Table~\ref{suptab:animate}, we show the result for the domain change (inanimate and animate) based split, again splitting the samples randomly or using clustering. In both the cases, our score fusion method performs better than other baselines thereby showing the generalizability of the approach to various practical scenrios.

\begin{table}[t!]
    \centering
    \caption{Disjoint-CIL analysis with deeper backbone ResNet18 (ResNet50 in Table~\ref{suptab:deepnet50}), 40 novel classes using random class split. Our observations generalize to deeper architectures. \ginanote{split2, 800-40-0, resnet18}}\label{suptab:deepnet18}
    \setlength{\tabcolsep}{5pt}
    \resizebox{\linewidth}{!}{
    \begin{tabular}{c|c|cc|c}
    \toprule
    Method & $Acc_{all}$ & $Acc_{base}$ & $Acc_{novel}$ & $Acc_{avg}$ \\ \midrule
    confidence-based routing                    & 49.91 & 47.98 & 88.50 & 68.24 \\
    learning-based routing w/ $L_{rt-bal}$   & 65.59	& 65.02	& 77.00	& 71.01 \\
    oracle routing w/ $L_{rt-bal}$           & 65.85 & 65.02	& 82.50	& 73.76 \\ \hline
    FA (ours) + BiC~\cite{bic}                           & 69.12 & 69.70	& 57.70	& 63.70 \\
    FA (ours) + WA~\cite{wa}                             & 68.88	& 69.02	& 66.10	& 67.56 \\ \midrule
score fusion (ours) best-$Acc_{all}$             & \textbf{69.45} & 70.01 & 58.13 & 64.07 \\
score fusion (ours) best-balanced                & 67.36 & 66.61 & 82.37 & 74.49 \\
score fusion (ours) best-$Acc_{avg}$             & 65.83 & 64.85 & 85.50 & \textbf{75.17} \\ \midrule
    joint learning (oracle)                              & 70.32 & 70.43	& 68.20 & 69.32 \\\bottomrule
    \end{tabular}}

\end{table}

\begin{table}[t!]
    \centering
    \caption{Disjoint-CIL analysis with deeper backbone ResNet50 (see Table~\ref{suptab:deepnet18} for ResNet18), 40 novel classes using random class split. Our observations generalize to these architectures. 
    }\label{suptab:deepnet50} %
    \setlength{\tabcolsep}{5pt}
    \resizebox{\linewidth}{!}{
    \begin{tabular}{c|c|cc|c}
    \toprule
    Method & $Acc_{all}$ & $Acc_{base}$ & $Acc_{novel}$ & $Acc_{avg}$ \\ \midrule
    confidence-based routing                    & 60.15 & 58.65 & 90.10 & 74.38 \\
    learning-based routing w/ $L_{rt-bal}$   & 73.21 & 72.83 & 80.97 & 76.90 \\ \midrule
score fusion (ours) best-$Acc_{all}$             & \textbf{75.40} & 75.94 & 64.57 & 70.25 \\
score fusion (ours) best-balanced                & 74.44 & 74.00 & 83.13 & 78.57 \\
score fusion (ours) best-$Acc_{avg}$             & 72.02 & 71.24 & 87.43 & \textbf{79.34} \\ \midrule
    joint learning (oracle)                              & 76.59	& 76.71 & 74.10 & 75.41 \\\bottomrule
    \end{tabular}}
\end{table}

\begin{table}[t!]
    \centering
    \caption{Disjoint-CIL analysis with ResNet10, larger novel set with 200 classes using random class splits. Our observations hold. 
    }
    \setlength{\tabcolsep}{5pt}
    \resizebox{1\linewidth}{!}{
    \begin{tabular}{c|c|cc|c}
    \toprule
    Method & $Acc_{all}$ & $Acc_{base}$ & $Acc_{novel}$ & $Acc_{avg}$ \\ \midrule
    confidence-based routing                    & 51.25 & 44.84 & 76.88 & 60.86 \\
    learning-based routing w/ $L_{rt-bal}$   & 56.91 & 54.39	& 67.01 & 60.70 \\
    oracle routing w/ $L_{rt-bal}$           & 57.80	& 54.95 & 69.18 & 62.07 \\ \hline
    FA (ours) + BiC~\cite{bic}                           & 61.09	& 61.62 & 58.98 & 60.30 \\
    FA (ours) + WA~\cite{wa}                             & 54.42	& 64.31 & 14.88 & 39.60 \\ \midrule
score fusion (ours) best-$Acc_{all}$             & \textbf{61.56} & 61.39 & 62.21 & 61.80 \\
score fusion (ours) best-balanced                & 61.29 & 59.63 & 67.93 & 63.78 \\
score fusion (ours) best-$Acc_{avg}$             & 59.57 & 56.31 & 72.60 & \textbf{64.45} \\ \midrule
    joint learning (oracle)                              & 61.68	& 61.42	& 62.76	& 62.09 \\\bottomrule
    \end{tabular}\label{suptab:disjoint200}
    }
\end{table}

\begin{table}[t!]
    \centering
    \caption{Overlapping-CIL results of random class splits.
    }\label{suptab:randcls}
    \setlength{\tabcolsep}{3pt}
    \begin{subtable}[t]{\linewidth}
    \caption{\footnotesize Overlapping classes samples split randomly.}
    \resizebox{1\linewidth}{!}{
    \begin{tabular}{c|c|ccc|c}
    \toprule
    Method & $Acc_{all}$ & $Acc_{base}$ & $Acc_{novel}$ & $Acc_{ovlp}$ & $Acc_{avg}$ \\ \midrule
    confidence-based routing                    & 39.91 & 37.67 & 84.80 & 83.20 & 68.56 \\
    learning-based routing w/ $L_{rt-bal}$   & 55.32 & 54.72 & 68.34 & 59.20 & 60.75 \\
    oracle routing w/ $L_{rt-bal}$           & 55.54 & 54.71 & 74.06 & 58.40 & 62.39 \\ \hline
    logit concatenation (avg pool)              & 59.74 & 59.95 & 56.00 & 51.20 & 55.72 \\
    logit concatenation (max pool)              & 59.65 & 59.80 & 55.89 & 60.80 & 58.83 \\ \midrule
score fusion (ours) best-$Acc_{all}$             & \textbf{59.88} & 60.40 & 48.61 & 55.47 & 54.83 \\
score fusion (ours) best-balanced                & 57.99 & 57.24 & 74.02 & 65.33 & 65.53 \\
score fusion (ours) best-$Acc_{avg}$             & 53.09 & 51.70 & 81.64 & 72.27 & \textbf{68.54} \\ \midrule
    joint learning (oracle)                              & 60.25 & 60.35 & 56.91 & 68.00 & 61.75 \\\bottomrule
    \end{tabular}}
    \end{subtable} \\ \hspace{0.5pt}
    \begin{subtable}[t]{\linewidth}
    \caption{\footnotesize Overlapping classes samples split by clustering.
    }
    \resizebox{1\linewidth}{!}{
    \begin{tabular}{c|c|ccc|c}
    \toprule
    Method & $Acc_{all}$ & $Acc_{base}$ & $Acc_{novel}$ & $Acc_{ovlp}$ & $Acc_{avg}$ \\ \midrule
    confidence-based routing                    & 37.37 & 35.26 & 79.43 & 80.00 & 64.90 \\
    learning-based routing w/ $L_{rt-bal}$   & 52.97 & 52.49 & 65.30	& 44.27 & 54.02 \\
    oracle routing w/ $L_{rt-bal}$           & 52.74 & 51.95 & 71.66 & 45.60 & 56.40 \\ \hline
    logit concatenation (avg pool)              & 56.41	& 56.57 & 54.86 & 42.40 & 51.28 \\
    logit concatenation (max pool)              & 56.34	& 56.35 & 54.86 & 64.80 & 58.67 \\ \midrule
score fusion (ours) best-$Acc_{all}$             & \textbf{57.00} & 57.41 & 48.11 & 54.13 & 53.22 \\
score fusion (ours) best-balanced                & 55.33 & 54.54 & 71.24 & 69.60 & 65.13 \\
score fusion (ours) best-$Acc_{avg}$             & 49.51 & 48.13 & 76.76 & 77.33 & \textbf{67.41} \\ \midrule
    joint learning (oracle)                              & 57.29	& 57.51 & 51.77 & 60.00 & 56.43 \\\bottomrule
    \end{tabular}}
    \end{subtable}
\end{table}

\begin{table}[t!]
    \centering
    \caption{Overlapping-CIL results of domain-changing (inanimate-animate) splits.}\label{suptab:animate}
    \setlength{\tabcolsep}{3pt}
    \begin{subtable}[t]{\linewidth}
    \caption{\footnotesize Overlapping classes samples split randomly.
    }
    \resizebox{1\linewidth}{!}{
    \begin{tabular}{c|c|ccc|c}
    \toprule
    Method & $Acc_{all}$ & $Acc_{base}$ & $Acc_{novel}$ & $Acc_{ovlp}$ & $Acc_{avg}$ \\ \midrule
    confidence-based routing                    & 40.82	& 37.75 & 83.43 & 91.20 & 70.79 \\
    learning-based routing w/ $L_{rt-bal}$   & 56.18 & 54.58 & 81.71 & 58.67 & 64.99 \\
    oracle routing w/ $L_{rt-bal}$           & 56.70	& 55.01 & 83.77 & 58.40 & 65.73 \\ \hline
    logit concatenation (avg pool)              & 57.34 & 56.25 & 74.17 & 64.00 & 64.81 \\
    logit concatenation (max pool)              & 55.84 & 54.55 & 74.06 & 75.20 & 67.94 \\ \midrule
score fusion (ours) best-$Acc_{all}$             & \textbf{57.44} & 56.27 & 75.81 & 61.87 & 64.65 \\
score fusion (ours) best-balanced                & 55.66 & 53.88 & 81.75 & 75.20 & 70.28 \\
score fusion (ours) best-$Acc_{avg}$             & 52.55 & 50.44 & 82.40 & 84.27 & \textbf{72.37} \\ \midrule
    joint learning (oracle)                              & 57.95	& 56.52 & 80.11 & 65.60 & 67.41 \\\bottomrule
    \end{tabular}}
    \end{subtable} \\ \hspace{0.5pt}
    \begin{subtable}[t]{\linewidth}
    \caption{\footnotesize Overlapping classes samples split by clustering.
    }
    \resizebox{1\linewidth}{!}{
    \begin{tabular}{c|c|ccc|c}
    \toprule
    Method & $Acc_{all}$ & $Acc_{base}$ & $Acc_{novel}$ & $Acc_{ovlp}$ & $Acc_{avg}$ \\ \midrule
    confidence-based routing                    & 38.28 & 35.30 & 80.23 & 82.40 & 65.98 \\
    learning-based routing w/ $L_{rt-bal}$   & 52.89 & 51.37 & 78.90 & 44.00 & 58.09 \\
    oracle routing w/ $L_{rt-bal}$           & 53.30	& 51.66 & 81.03 & 45.60 & 59.43 \\ \hline
    logit concatenation (avg pool)              & 53.57 & 52.61 & 70.17 & 46.40 & 56.39 \\
    logit concatenation (max pool)              & 52.34 & 51.06 & 70.17 & 72.00 & 64.41 \\ \midrule
score fusion (ours) best-$Acc_{all}$             & \textbf{54.11} & 52.84 & 74.40 & 56.80 & 61.35 \\
score fusion (ours) best-balanced                & 52.84 & 51.13 & 78.06 & 69.87 & 66.35 \\
score fusion (ours) best-$Acc_{avg}$             & 51.55 & 49.73 & 77.94 & 73.33 & \textbf{67.00} \\ \midrule
    joint learning (oracle)                              & 54.28	& 52.70	& 78.06	& 67.20	& 65.99 \\\bottomrule
    \end{tabular}}
    \end{subtable}
\end{table}
 \section{Implementation Details}
 
In all the experiments, we adopt batch size 256 with learning rate starts at 0.1, and normalize the features and weights. The pretrained model is trained for 90 epochs with the SGD optimizer, where the learning rate decays every 30 epochs. In the first stage, we finetune the network for 30 epochs with learning rate decays every 10 epochs. As mentioned in section~\ref{sec:fusion} in the main paper, The second-stage training adopts class-balanced sampling, which is trained for 10 epochs. Experiments are repeated with different random seeds.

\end{document}